\documentclass[letterpaper]{article} 
\usepackage{aaai2026}  
\usepackage{times}  
\usepackage{helvet}  
\usepackage{amsmath}
\usepackage{amssymb}
\usepackage{pdflscape}
\usepackage{bm}
\usepackage{threeparttable}
\usepackage{subcaption}
\usepackage{booktabs}
\usepackage{multirow}
\usepackage{graphicx}
\usepackage{courier}  
\usepackage[hyphens]{url}  
\usepackage{graphicx} 
\usepackage{natbib}  
\usepackage{caption} 
\usepackage{algorithm}
\usepackage{algorithmic}

\usepackage{newfloat}
\usepackage{listings}
\DeclareCaptionStyle{ruled}{labelfont=normalfont,labelsep=colon,strut=off} 
\floatstyle{ruled}
\newfloat{listing}{tb}{lst}{}
\floatname{listing}{Listing}
\title{Flexible Concept Bottleneck Model}

\author {
    Xingbo Du\textsuperscript{\rm 1}$^{\dag *}$,
    Qiantong Dou\textsuperscript{\rm 2}$^\dag$,
    Lei Fan\textsuperscript{\rm 2},
    Rui Zhang\textsuperscript{\rm 3}\thanks{Corresponding authors, $^\dag$Equal contribution.}}
\affiliations {
    \textsuperscript{\rm 1}Mohamed bin Zayed University of Artificial Intelligence\\
    \textsuperscript{\rm 2}School of Computer Science and Engineering, University of New South Wales\\
    \textsuperscript{\rm 3}Gaoling School of Artificial Intelligence, Renmin University of China\\
Xingbo.Du@mbzuai.ac.ae, q.dou@student.unsw.edu.au, lei.fan1@unsw.edu.au, rayzhang@ruc.edu.cn

}

\usepackage{bibentry}

\begin{document}

\maketitle

\begin{abstract}
Concept bottleneck models (CBMs) improve neural network interpretability by introducing an intermediate layer that maps human-understandable concepts to predictions. Recent work has explored the use of vision-language models (VLMs) to automate concept selection and annotation. However, existing VLM-based CBMs typically require full model retraining when new concepts are involved, which limits their adaptability and flexibility in real-world scenarios, especially considering the rapid evolution of vision-language foundation models.
To address these issues, we propose Flexible Concept Bottleneck Model (FCBM), which supports dynamic concept adaptation, including complete replacement of the original concept set. Specifically, we design a hypernetwork that generates prediction weights based on concept embeddings, allowing seamless integration of new concepts without retraining the entire model. In addition, we introduce a modified sparsemax module with a learnable temperature parameter that dynamically selects the most relevant concepts, enabling the model to focus on the most informative features. Extensive experiments on five public benchmarks demonstrate that our method achieves accuracy comparable to state-of-the-art baselines with a similar number of effective concepts. Moreover, the model generalizes well to unseen concepts with just a single epoch of fine-tuning, demonstrating its strong adaptability and flexibility.
\end{abstract}

\begin{links}
    \link{Code}{https://github.com/deepopo/FCBM}
\end{links}

\section{Introduction}
Understanding black-box models is a core challenge in developing trustworthy and interpretable systems for fairness~\citep{mehrabi2021survey}, value alignment~\citep{gabriel2020artificial}, and human–machine interaction~\citep{zhang2022rethinking}. To better understand how deep models work internally, researchers \citep{dwivedi2023explainable} have proposed a range of interpretability methods, such as case-based reasoning~\citep{li2018deep,zhu2025interpretable} and disentangled neural networks~\citep{kim2018interpretability}. Among them, concept bottleneck models (CBMs)~\citep{koh2020concept} enhance model transparency by explicitly introducing an intermediate concept layer, where the model first predicts human-interpretable concepts before making the final task prediction. Given training data annotated by experts, a standard CBM architecture consists of two components: a concept predictor, which maps the input into a bottleneck layer where each unit represents a human-understandable concept; and a linear classifier, which uses these predicted concepts to produce the final output. By disentangling high-level concepts from raw inputs, CBMs allow users to understand and directly intervene in the model's decision process~\citep{chauhan2023interactive}.

CBMs require large-scale expert-labeled concepts to cover a sufficient range of data features, resulting in substantial costs and limited scalability. To address this issue, recent studies~\citep{yuksekgonul2022post, oikarinen2023labelfree, yang2023language,tang2025prototype} use pre-trained vision–language models (VLMs) to eliminate the need for manual annotation. Specifically, a VLM-based CBM first uses a large language model (LLM)~\cite{brown2020language} to generate a pool of candidate concepts, and then employs foundation models (\textit{e.g.}, CLIP~\citep{radford2021learning}) to compute similarity scores between inputs and each concept in the pool as concept values. This approach enables training without expert annotations and allows any black-box classifier to be converted into a transparent white-box model. 

VLM-based CBMs use the powerful alignment capabilities of VLM to automatically construct concept pools from images, enhancing the scalability of concept acquisition. However, these concept sets are typically fixed during both training and inference, which limits the model’s flexibility and adaptability to new contexts. In many real-world scenarios, such as medicine, newly discovered biomarkers need to be incorporated into the concept space. Likewise, the rapid iteration of vision-language foundation models often shifts the underlying semantic representations, necessitating updates to the concept embeddings. Therefore, the fixed concept pool must be redefined and re-aligned, requiring costly end-to-end retraining. This rigidity severely constrains the practical deployment of existing CBMs, particularly in real-world applications where domain knowledge evolves continuously and foundation models are frequently updated~\citep{zhang2025way}.

\begin{figure}[t]
\centering
\includegraphics[width=0.98\columnwidth]{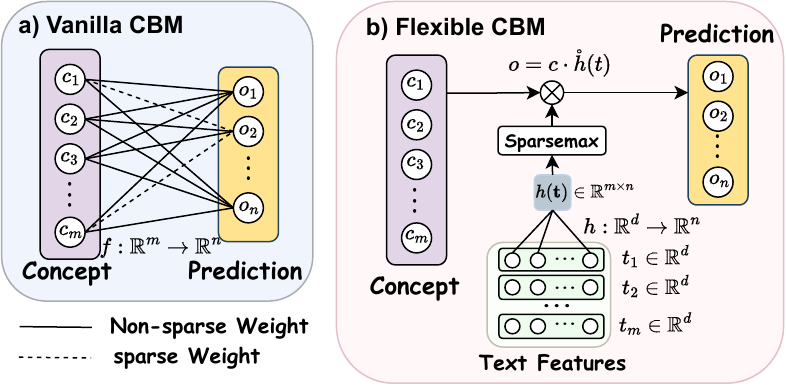}
\caption{Difference between the vanilla CBM and FCBM.}
\label{fig1}
\end{figure}

To address these challenges, we propose Flexible Concept Bottleneck Model (FCBM), which extends existing VLM-based CBMs by allowing the concept pool to be modified dynamically, as illustrated in Fig.~\ref{fig1}. Unlike previous CBM variants that rely on a fixed concept-to-label mapping, FCBM employs a hypernetwork~\citep{ha2016hypernetworks} to disentangle concept representations from their contribution weights. Equipped with the hypernetwork, FCBM incorporates new concepts and generates adapted weights dynamically, thereby supporting the integration of evolving knowledge. To maintain interpretability and mitigate concept dilution, we introduce a sparsity strategy~\citep{martins2016softmax} to recalibrate the balance between prediction accuracy and concept activation. This encourages sparse explanations, ensuring that only a concise set of key concepts is activated for each prediction. \textbf{The main contributions are as follows}:

\noindent 1) \textbf{Flexible and scalable framework.} We propose a novel framework named \textbf{F}lexible \textbf{C}oncept \textbf{B}ottleneck \textbf{M}odel (FCBM). Unlike existing approaches using a fixed concept pool, FCBM adapts to evolving knowledge when involving new concepts, which is more scalable and adaptable in scenarios where different concept subsets are preferable or when users want to leverage more advanced LLMs.

\noindent 2) \textbf{Dynamic concept modeling with sparse and interpretable selection.} To enable flexible concept usage, FCBM introduces a hypernetwork to generate weights dynamically for new or updated concepts. In addition, we design a sparsemax module with a learnable temperature, which promotes sparsity in concept selection and ensures that only the most informative concepts are activated, preserving interpretability while avoiding concept redundancy.

\noindent 3) \textbf{Comparable accuracy to state-of-the-art baselines.} We execute extensive experiments to show that FCBM is comparable to state-of-the-art baselines with two backbones and outperforms all baselines on more than half of the benchmarks. Moreover, the model generalizes well to unseen concepts with just a single epoch of fine-tuning, demonstrating its strong adaptability and flexibility.

\section{Related Work}

\textbf{Concept Bottleneck Models (CBMs)}. As deep neural networks are increasingly deployed, understanding their internal decision processes has become a key challenge~\citep{doshi2017towards}. A common approach is to align internal representations with human-interpretable concepts~\citep{kim2018interpretability}, promoting transparency and trust. While post-hoc methods offer several insights, they often fail to reflect the model’s true reasoning~\citep{rudin2019stop}. This motivated a shift towards methods that embed interpretability directly into the model by design, among which one prominent design is CBM~\citep{koh2020concept}. CBMs~\citep{koh2020concept,losch2021semantic} and their variants~\citep{kazhdan2020now,kim2023probabilistic} follow a two-stage architecture: they first map input images to a set of expert-defined concepts, and then perform classification based on these concepts via a linear layer. This structure allows predictions to be explained through human-understandable concepts and provides strong interactivity. Beyond interpretable image classification, CBMs have been applied to diverse areas including human–AI interaction~\citep{chauhan2023interactive}, decision-making processes~\citep{das2023state2explanation}, and causal inference~\citep{dominici2025causal}. However, training CBMs requires large expert-labeled concepts, which are costly to obtain. Theoretically, under limited expert supervision, CBMs face a trade-off between interpretability and predictive accuracy~\citep{zhang2024decoupling}. Although several methods have been proposed to mitigate this trade-off, such as CBM-AUC~\citep{sawada2022concept}, DCBM~\citep{zhang2024decoupling}, and CEM~\citep{espinosa2022concept}, they still rely on concept-level supervision and are not applicable in scenarios where expert-labeled concepts are unavailable or extremely limited.

\begin{figure*}[t]
\centering
\includegraphics[width=0.8\textwidth]{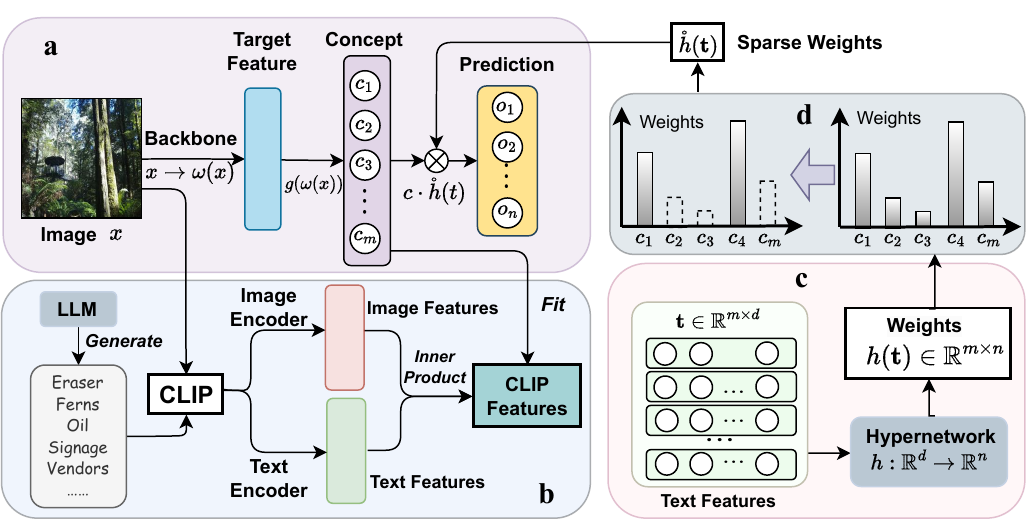} 
\caption{The pipeline of FCBM consists of four key components: a) A two-stage learning framework; b) Concept sets generated by LLMs, which are used to form CLIP-derived features; c) A hypernetwork that generates weights based on text features; and d) A tailored sparsemax module that enforces sparsity in the weights during both training and inference.}
\label{fig:pipeline}
\end{figure*}

\textbf{VLM-based CBMs.} Recent studies have also explored constructing CBMs by leveraging the rich knowledge reserve of LLMs~\citep{berrios2023towards} and the semantic understanding capabilities of VLMs~\citep{wang2024sam,zang2025sage}, to reduce the reliance on concept annotations. The Post-hoc CBM~\citep{yuksekgonul2022post} leverages external knowledge such as ConceptNet~\citep{speer2017conceptnet} to identify meaningful concepts and align them with input features through CLIP~\citep{radford2021learning}. Label-free CBM~\citep{oikarinen2023labelfree} first generates the concept set by prompting GPT-3~\citep{brown2020language}, and the mapping from image embeddings to concept values in the CLIP space is optimized using a cosine cubed loss. In contrast, Labo~\citep{yang2023language} uses submodular optimization~\citep{bach2010convex} to filter candidate concepts, ensuring the discriminative representations and semantic diversity. Recently, \citeauthor{panousis2024coarse} proposed a hierarchical reasoning strategy to improve the predictive performance and interoperability. VLG-CBM~\citep{srivastava2024vlg} leverages an open-domain grounded object detection model~\citep{liu2024grounding} to generate localized and visually interpretable concept annotations. By filtering out non-visual concepts using visual signals from open-vocabulary object detectors, it enhances both transparency and explainability while improving prediction performance. In addition, OpenCBM~\citep{tan2024explain} aims to support flexible, open-vocabulary concepts specified by users at test time. While offering a degree of flexibility, its capabilities are limited to adding or removing concepts, and it cannot accommodate complete updates to the concept pool, a requirement in scenarios where foundation models are updated.

\section{Methodology}

Our proposed framework FCBM is illustrated in Fig.~\ref{fig:pipeline}, which consists of four main components: 1) a two-stage learning framework (Sec.~\ref{sec:framework}), 2) concept set generation via LLMs and CLIP feature formation (Sec.~\ref{sec:concept_generate}), 3) a hypernetwork for dynamic concept-to-label mapping (Sec.~\ref{sec:hypernet}), and 4) a tailored sparsemax module for interpretable concept selection (Sec.~\ref{sec:sparsemax}).

\subsection{Preliminaries and Framework of FCBM}\label{sec:framework}
In Concept Bottleneck Models (CBMs), three core components are involved: input images, concepts, and labels, which are represented as vectors $\bm{x_i}$, $\bm{c_i}$, and $\bm{y_i}$ ($i=1,2,\cdots,N$), respectively. CBMs typically follow a two-stage learning framework: a concept predictor $g: \mathcal{X} \rightarrow \mathcal{C}$ that maps images $\bm{x}$ to concepts $\bm{c}$, followed by a label predictor $f: \mathcal{C} \rightarrow \mathcal{Y}$ that maps concepts $\bm{c}$ to final predictions $\bm{y}$.

In vanilla CBMs~\citep{koh2020concept}, ground-truth concepts $\bm{c}$ are assumed to be annotated by human experts, which is both time-consuming and challenging. To address this issue, VLM-based CBMs~\citep{oikarinen2023labelfree} automatically generate concepts by combining capabilities of VLMs and LLMs (see Sec.~\ref{sec:concept_generate}). By eliminating the dependence on expert-annotated concepts, CBMs can be trained on a much broader range of datasets, such as Places365~\cite{zhou2017places} and ImageNet~\cite{deng2009imagenet}.

FCBM also follows a two-stage learning framework but introduces several design differences. Specifically, given images $\bm{v}$, a backbone model $\omega$ is used to extract target features $\omega(\bm{x})$. Let $\bm{q} \triangleq g \circ \omega(\bm{x})$, then the first stage optimizes the concept predictor by minimizing the following objective:
\begin{equation}\label{eq:x-c}
{g^*} = \underset{g}{\arg\min} \sum_{j=1}^{m} \left[-\text{sim}({\bm{c}}_{:,j}, \bm{q}_{:, j})\right],
\end{equation}
where $m$ is the number of concepts, $\text{sim}(\cdot, \cdot)$ is the \textit{cosine-cubed} similarity introduced in LF-CBM~\citep{oikarinen2023labelfree}, and ${\bm{c}}$ represents the CLIP-derived features (Sec.~\ref{sec:concept_generate}). 

In most VLM-based CBMs~\citep{oikarinen2023labelfree,srivastava2024vlg}, the second stage simply aims to minimize the Cross Entropy (CE) between the predicted outcome $f \circ {g^*} \circ \omega(\bm{x})$ and labels $\bm{y}$, where $f: \mathcal{C}\rightarrow\mathcal{Y}$ is a linear-like projection. However, this formulation restricts $f$ to a fixed number of concepts in $\mathcal{C}$, which is not well-suited for dynamic concepts. In contrast, FCBM introduces a hypernetwork $h$ that generates weights $h(\bm{t})$ from text features $\bm{t}$ and uses a tailored sparsemax operation $\mathcal{S}_{\max}^{\tau}$ with temperature $\tau$ to obtain the sparse weights $\mathring{h}(\bm{t})\triangleq \mathcal{S}_{\max}^{\tau}(h(\bm{t}))$.

Thus, the second learning stage can be written as:
\begin{equation}\label{eq:c-y}
{h^*} = \underset{{h}}{\arg\min}\ \mathcal{L},\quad \mathcal{L} = \sum_{i=1}^N \text{CE}({g^*} \circ \omega(\bm{x_i}) \cdot \mathring{h}(\bm{t}), \bm{y_i}),
\end{equation}
where the $\mathring{h}^*(\bm{t})$ is used for the classification tasks.

\subsection{Concept Generation and CLIP Features}\label{sec:concept_generate}
Concepts are generated through large language models (LLMs), utilizing the same prompt templates introduced in LF-CBM~\citep{oikarinen2023labelfree}, where GPT-3~\citep{brown2020language} serves as the base model. To ensure comparability, we adopt the same generated concepts for the main experiments. Additionally, we include concepts generated by more advanced LLMs, such as DeepSeek-V3~\citep{liu2024deepseek} and GPT-4o~\citep{openai2024gpt4o}, to demonstrate the zero-shot generalization capability of FCBM with respect to concepts.

A vision-language model (\textit{e.g.}, CLIP~\citep{radford2021learning}) is then employed as both an image encoder and a text encoder to extract image features $\bm{z} \in \mathbb{R}^{N \times d}$ from images and text features $\bm{t} \in \mathbb{R}^{m \times d}$ from concept texts, where $N$, $m$, and $d$ represent the number of images, the number of concepts, and the feature dimensions, respectively. Both modalities are encoded into a shared feature space of dimension $d$, as they are pretrained through a contrastive learning mechanism~\citep{desai2021virtex}. As a result, when a concept text accurately describes an image's characteristics, their corresponding features exhibit high similarity in the joint embedding space.

The CLIP-derived features are computed by taking the inner product between $\bm{z}$ and $\bm{t}$, forming $\bm{c} = \bm{z} \cdot \bm{t}^\top \in \mathbb{R}^{N \times m}$. Here, each element $\bm{c_i} \in \mathbb{R}^{m}$ represents the concept vector for the image $\bm{x_i}$.

\subsection{Hypernetwork}\label{sec:hypernet}
To better extract informative details from concepts and support a dynamic number of concepts, we introduce a hypernetwork $h: \mathbb{R}^{d} \rightarrow \mathbb{R}^{n}$, which maps the text feature dimension to the class dimension, where $n$ represents the number of classes. Notably, the scale of $h$ is independent of the number of concepts $m$, making it scalable to varying concept set sizes. The output of hypernetwork $h(\bm{t}) \in \mathbb{R}^{m \times n}$ shares the same shape as the weight matrix used in the linear projection $f: \mathcal{C} \rightarrow \mathcal{Y}$ in vanilla VLM-based CBMs~\citep{oikarinen2023labelfree}. Intuitively, $h(\bm{t})$ also acts as the weights, determining the contribution of each concept to the final predictions.

To preserve the zero-shot generalization capability with respect to concepts, we align the feature distributions during training and inference. Specifically, given the text features $\bm{t} \in \mathbb{R}^{m \times d}$ during training, we compute their mean $\bar{\bm{t}} \in \mathbb{R}^{d}$ and standard deviation $\sigma_{\bm{t}} \in \mathbb{R}^{d}$, as well as the mean $\bar{h}(\bm{t})\in \mathbb{R}^n$ and standard deviation $\sigma_{h(\bm{t})}\in \mathbb{R}^n$ of the hypernetwork's output $h(\bm{t}) \in \mathbb{R}^{m\times n}$. During inference, given the new text features $\bm{t}^{\prime}$ generated from a new set of concepts, the weights $\tilde{h}({\bm{t}}^{\prime})$ are obtained as follows:
\begin{equation}\label{eq:hypernet}
    \tilde{\bm{t}}^{\prime} \triangleq \frac{\sigma_{\bm{t}}}{\sigma_{\bm{t}^{\prime}}} (\bm{t}^{\prime} - \overline{\bm{t}^{\prime}}) + \bar{\bm{t}},\quad 
    \tilde{h}({\bm{t}}^{\prime}) \triangleq \frac{\sigma_{h(\bm{t})}}{\sigma_{h(\tilde{\bm{t}}^{\prime})}}(h(\tilde{\bm{t}}^{\prime}) - {\bar{h}(\tilde{\bm{t}}^{\prime}}))+\bar{h}(\bm{t}), 
\end{equation}
where $\overline{\bm{t}^{\prime}}$ and $\sigma_{\bm{t}^{\prime}}$ are the mean and standard deviation of $\bm{t}^{\prime}$, and $\bar{h}(\tilde{\bm{t}}^{\prime})$ and $\sigma_{h(\tilde{\bm{t}}^{\prime})}$ are the mean and standard deviation of $h(\tilde{\bm{t}}^{\prime})$. Intuitively, Eq.~\ref{eq:hypernet} ensures that the distribution of the final weights $\tilde{h}({\bm{t}}^{\prime})$ is aligned with the distribution of $h(\bm{t})$ in both the text feature dimension and the weight dimension.

\subsection{Sparsemax with a Learnable Temperature}\label{sec:sparsemax}

\begin{algorithm}[tb]
\caption{Sparsemax with temperature $\tau$ ($\mathcal{S}_{\max}^{\tau}$)}
\label{alg:sparsemax}
\textbf{Input}: Input vector $\bm{s}$, temperature $\tau$.\\
\makebox[\linewidth][l]{\textbf{Output}: $\tilde{\bm{s}}$, where $\tilde{\bm{s}}_{i} = [\bm{s}_i - \xi(\bm{s})]_+$.}
\begin{algorithmic}[1] 
\STATE Sort $\bm{s}$: $\bm{s}_{(1)} \geq \bm{s}_{(2)} \cdots \geq \bm{s}_{(m)}$.
\STATE Find $k(\bm{s})\triangleq \max\left\{k\in [m] | \tau + k \bm{s}_{(k)} > \sum_{j\leq k} \bm{s}_{(j)}\right\}$.
\STATE Let $\xi(\bm{s})=\frac{\left(\sum_{j\leq k(\bm{s})} \bm{s}_{(j)}\right) - \tau}{k(\bm{s})}$.
\end{algorithmic}
\end{algorithm}

The weights $h(\bm{t})$ generated by the hypernetwork are generally non-sparse, which may hinder interpretability. To address this issue, we introduce a modified sparsemax module based on~\citep{martins2016softmax}, further incorporating a learnable temperature parameter $\tau$ to dynamically control the level of sparsity. This module can enforce sparsity in the weight selection by controlling the number of active concepts used for each prediction.
Intuitively, sparsemax acts similarly to softmax but generates sparse outputs, enabling the model to focus on the most relevant concepts. The temperature parameter $\tau$ adjusts the level of sparsity: higher temperatures lead to fewer active concepts, while lower temperatures increase the number of concepts considered in the prediction. By making $\tau$ learnable in the training stage, the model can automatically balance prediction accuracy and interpretability. The overall algorithm of the modified sparsemax is shown in Alg.~\ref{alg:sparsemax}.

To optimize the temperature parameter jointly with other model parameters, we need to compute the gradient of the sparsemax operation with respect to both $\bm{s}$ and $\tau$. The original sparsemax optimization was proposed in \citep{martins2016softmax}, where the Jacobian of the transformation given a vector $\bm{v}$ is expressed as:
\begin{equation}\label{eq:jacabian}
J_{\text{sparsemax}}(\bm{s}) \cdot \bm{v} = \bm{e} \odot (\bm{v} - \hat{v}\mathbf{1}), \quad \text{with} \quad \hat{v} := \frac{\sum_{j \in P(\bm{s})} v_j}{|P(\bm{s})|}, 
\end{equation}
where $P(\bm{s})$ represents the support set (\textit{i.e.}, the indices where $\mathcal{S}_{\max}^{\tau}(\bm{s})_i > 0$), $\odot$ denotes the Hadamard (element-wise) product, and $\mathbf{1}$ is an all-ones vector. $\bm{e}$ is an indicator vector with $\bm{s_i} = 1$ if $i \in P(\bm{s})$ and $0$ otherwise. This computation is efficient, requiring only $O(|P(\bm{s})|)$ operations, especially when the support set is small.
The derivation of $\tau$ is:
\begin{equation}\label{eq:approximation}
\frac{\partial \mathcal{L}}{\partial \tau} = \sum_{i \in P(\mathbf{\bm{s}})} \frac{1}{|P(\bm{s})|}\cdot \frac{\partial \mathcal{L}}{\partial \tilde{\bm{s}}_i}.
\end{equation}

Detailed derivations of Eq.~\ref{eq:jacabian} and~\ref{eq:approximation} are in the Appendix.

\section{Experiments}

\begin{table*}[tb!]
\centering
\setlength{\tabcolsep}{4pt}
\small
\begin{tabular}{c|c|c|ccccc}
\toprule
\multirow{2}{*}[-0.25em]{\textbf{Backbone}} & 
\multirow{2}{*}[-0.25em]{\textbf{Method}} & 
\multirow{2}{*}[-0.25em]{\shortstack{\textbf{Average}\\\textbf{NEC}}} & 
\multicolumn{5}{c}{\textbf{Dataset}} \\ \cmidrule{4-8}
& & & CIFAR10 & CIFAR100 & CUB & Places365 & ImageNet \\ \midrule\midrule

\multirow{6}{*}[-0.25em]{ResNet50} 
& Standard & - & 88.55 {\scriptsize $\pm$ 0.02} & 70.19 {\scriptsize $\pm$ 0.06} & 71.00 {\scriptsize $\pm$ 0.16} & 53.28 {\scriptsize $\pm$ 0.04} & 73.14 {\scriptsize $\pm$ 0.04} \\ \cmidrule{2-8}
& Standard (sparse) & 29.84 & 82.11 {\scriptsize $\pm$ 0.00} & 57.54 {\scriptsize $\pm$ 0.01} & 53.34 {\scriptsize $\pm$ 0.01} & 44.00 {\scriptsize $\pm$ 0.00} & 57.03 {\scriptsize $\pm$ 0.01} \\
& PCBM & 32.37 & 76.43 {\scriptsize $\pm$ 0.02} & 56.24 {\scriptsize $\pm$ 0.06} & 58.39 {\scriptsize $\pm$ 0.06} & 42.26 {\scriptsize $\pm$ 0.35} & 61.38 {\scriptsize $\pm$ 0.22} \\
& LF-CBM & 27.37 & \textbf{86.16 {\scriptsize $\pm$ 0.05}} & \underline{64.62 {\scriptsize $\pm$ 0.06}} & 56.91 {\scriptsize $\pm$ 0.14} & \underline{48.88 {\scriptsize $\pm$ 0.05}} & \underline{66.03 {\scriptsize $\pm$ 0.05}} \\
& CF-CBM & 28.46 & 85.42 {\scriptsize $\pm$ 0.07} & 64.31 {\scriptsize $\pm$ 0.25} & \textbf{64.23 {\scriptsize $\pm$ 0.29}} & 46.39 {\scriptsize $\pm$ 0.34} & 65.95 {\scriptsize $\pm$ 0.27} \\
& FCBM (ours) & 28.85 & \underline{85.59 {\scriptsize $\pm$ 0.29}} & \textbf{64.77 {\scriptsize $\pm$ 0.41}} & \underline{63.46 {\scriptsize $\pm$ 0.41}} & \textbf{49.13 {\scriptsize $\pm$ 0.67}} & \textbf{66.34 {\scriptsize $\pm$ 0.34}} \\ \midrule

\multirow{6}{*}[-0.25em]{ViT-L/14}
& Standard & - & 98.02 {\scriptsize $\pm$ 0.03} & 86.99 {\scriptsize $\pm$ 0.09} & 85.22 {\scriptsize $\pm$ 0.05} & 55.66 {\scriptsize $\pm$ 0.06} & 84.11 {\scriptsize $\pm$ 0.05} \\ \cmidrule{2-8}
& Standard (sparse) & 28.92 & 96.61 {\scriptsize $\pm$ 0.00} & 80.28 {\scriptsize $\pm$ 0.00} & 72.07 {\scriptsize $\pm$ 0.00} & 44.90 {\scriptsize $\pm$ 0.01} & 72.96 {\scriptsize $\pm$ 0.13} \\
& PCBM & 32.56 & 92.13 {\scriptsize $\pm$ 0.05} & 74.29 {\scriptsize $\pm$ 0.13} & 68.25 {\scriptsize $\pm$ 0.32} & 43.18 {\scriptsize $\pm$ 0.36} & 65.79 {\scriptsize $\pm$ 0.53} \\
& LF-CBM & 27.85 & \underline{97.18 {\scriptsize $\pm$ 0.01}} & 81.98 {\scriptsize $\pm$ 0.02} & 75.44 {\scriptsize $\pm$ 0.04} & \underline{50.51 {\scriptsize $\pm$ 0.02}} & \underline{79.70 {\scriptsize $\pm$ 0.03}} \\
& CF-CBM & 28.56 & 96.35 {\scriptsize $\pm$ 0.05} & \underline{82.33 {\scriptsize $\pm$ 0.08}} & \underline{79.56 {\scriptsize $\pm$ 0.37}} & 48.55 {\scriptsize $\pm$ 0.28} & 79.16 {\scriptsize $\pm$ 0.44} \\
& FCBM (ours) & 28.08 & \textbf{97.21 {\scriptsize $\pm$ 0.06}} & \textbf{83.63 {\scriptsize $\pm$ 0.20}} & \textbf{80.52 {\scriptsize $\pm$ 0.27}} & \textbf{51.39 {\scriptsize $\pm$ 0.11}} & \textbf{80.62 {\scriptsize $\pm$ 0.32}} \\

\bottomrule
\end{tabular}
\caption{\label{tab:main}We compare prediction accuracy across five datasets using two backbones: ResNet50 and ViT-L/14. The average NEC is controlled to approximately 30, except for the \textit{Standard} model, to ensure a fair comparison. The best and the second best results, excluding the \textit{Standard} model, for each dataset and backbone are shown in \textbf{bold} and \underline{underline}, respectively.}
\end{table*}

The experimental protocols are outlined in Sec.~\ref{sec:exp_protocols}. Sec.~\ref{sec:exp_main} presents the main results, comparing results under a similar average number of effective concepts (NEC). To show the ability of FCBM's concept generalization, Sec.~\ref{sec:zero-shot} shows the zero-shot prediction accuracy using newly-generated concepts. Sec.~\ref{sec:ablation} provides ablation studies.

\subsection{Experimental Protocols}\label{sec:exp_protocols}

\paragraph{Datasets.} The experiments use five public datasets: CIFAR10~\cite{krizhevsky2009cifar}, CIFAR100~\cite{krizhevsky2009cifar}, CUB~\cite{wah2011caltech}, Places365~\cite{zhou2017places}, and ImageNet~\cite{deng2009imagenet}, consistent with previous works~\citep{oikarinen2023labelfree,zhang2024decoupling}. CIFAR10 and CIFAR100 are common image classification datasets containing 50,000 images with 10 and 100 classes, respectively. CUB is a fine-grained bird-species classification dataset with 4,900 training images and 200 classes. Places365 and ImageNet are large datasets with 1-2 million training images and 365 and 1,000 classes, respectively. The concepts used by FCBM are generated by LLMs using three types of prompts introduced in~\citep{oikarinen2023labelfree}: including the key features of a class, features commonly associated with a class, and the superclasses of a class.

\paragraph{Implementation Details.} 

Experiments were run using an AMD EPYC 7402 24-Core Processor, an NVIDIA GeForce RTX 4090, and 512GB RAM. The batch size was set to 50,000 for each dataset, with the Adam~\citep{kingma2015adam} optimizer and a learning rate of 0.001 for both objectives in Eq.\ref{eq:x-c} and Eq.\ref{eq:c-y}. The maximum number of epochs was set to 5,000 for the three small-scale datasets (CIFAR10, CIFAR100, and CUB), and 500 for the two large-scale datasets (ImageNet and Places365). This setting was empirically determined through a grid search.

\begin{figure*}[tb!]
\centering

\begin{subfigure}[t]{0.19\textwidth}
    \centering
    \includegraphics[width=\linewidth]{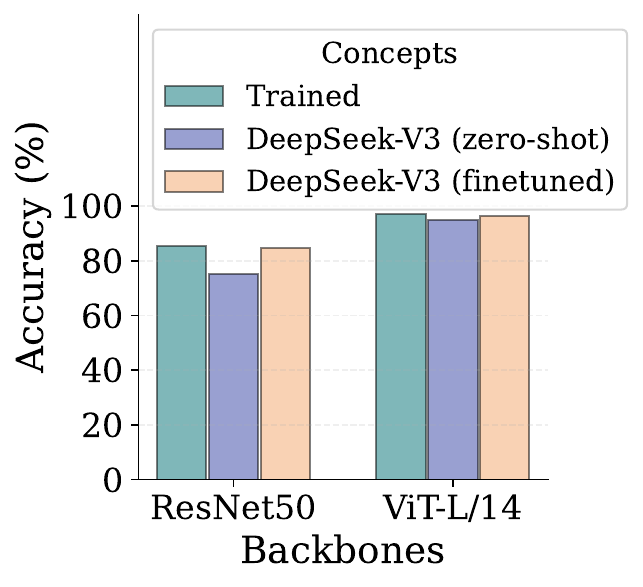}
    \caption{DeepSeek, CIFAR10}
\end{subfigure}
\hfill
\begin{subfigure}[t]{0.19\textwidth}
    \centering
    \includegraphics[width=\linewidth]{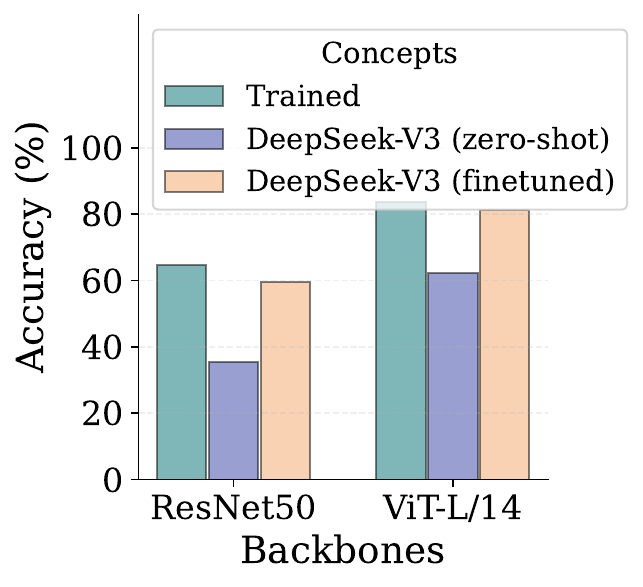}
    \caption{DeepSeek, CIFAR100}
\end{subfigure}
\hfill
\begin{subfigure}[t]{0.19\textwidth}
    \centering
    \includegraphics[width=\linewidth]{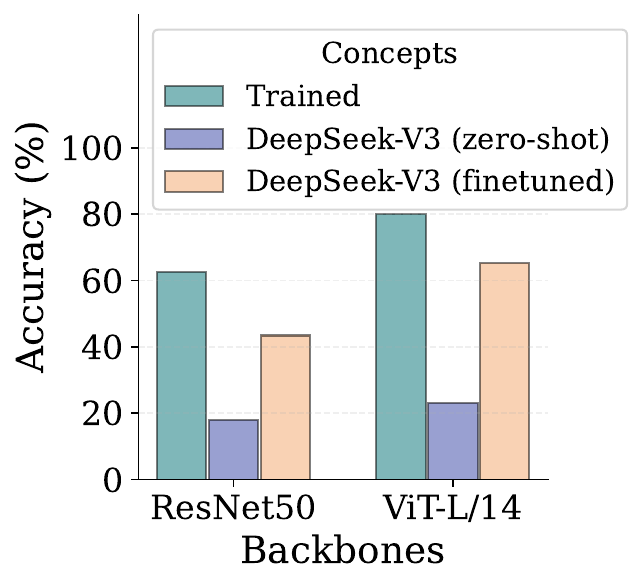}
    \caption{DeepSeek, CUB}
\end{subfigure}
\hfill
\begin{subfigure}[t]{0.19\textwidth}
    \centering
    \includegraphics[width=\linewidth]{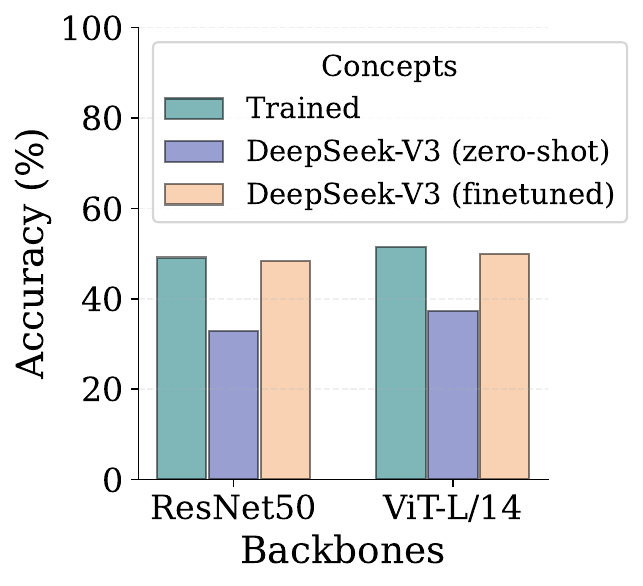}
    \caption{DeepSeek, Places365}
\end{subfigure}
\hfill
\begin{subfigure}[t]{0.19\textwidth}
    \centering
    \includegraphics[width=\linewidth]{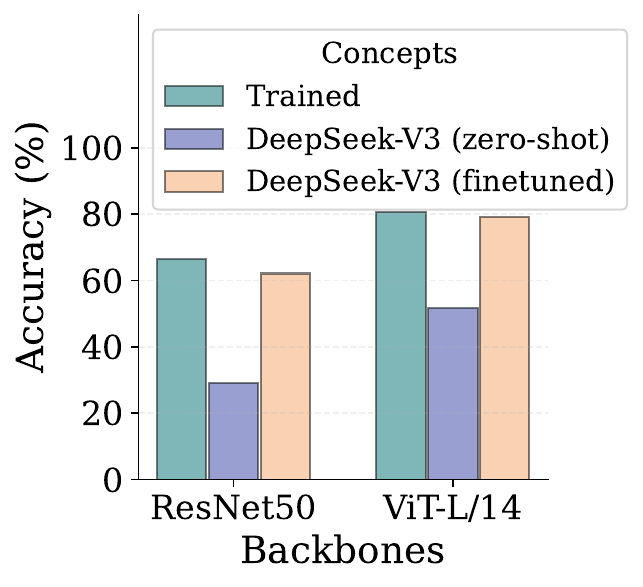}
    \caption{DeepSeek, ImageNet}
\end{subfigure}
\begin{subfigure}[t]{0.19\textwidth}
    \centering
    \includegraphics[width=\linewidth]{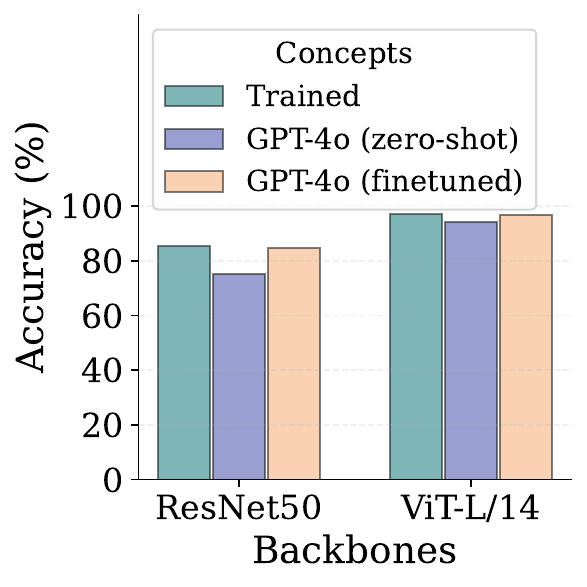}
    \caption{GPT, CIFAR10}
\end{subfigure}
\hfill
\begin{subfigure}[t]{0.19\textwidth}
    \centering
    \includegraphics[width=\linewidth]{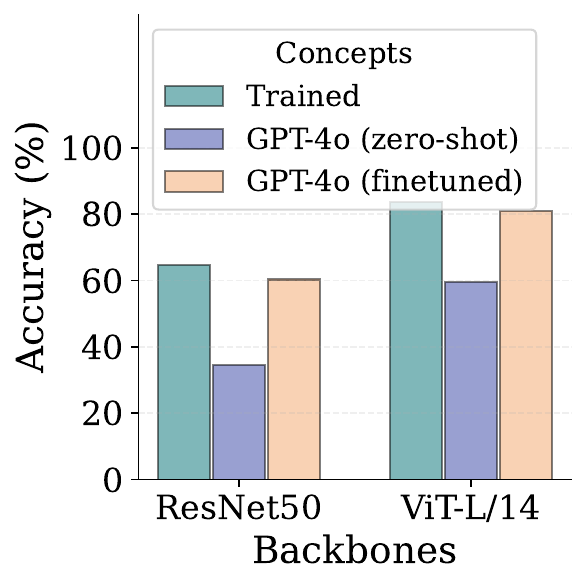}
    \caption{GPT, CIFAR100}
\end{subfigure}
\hfill
\begin{subfigure}[t]{0.19\textwidth}
    \centering
    \includegraphics[width=\linewidth]{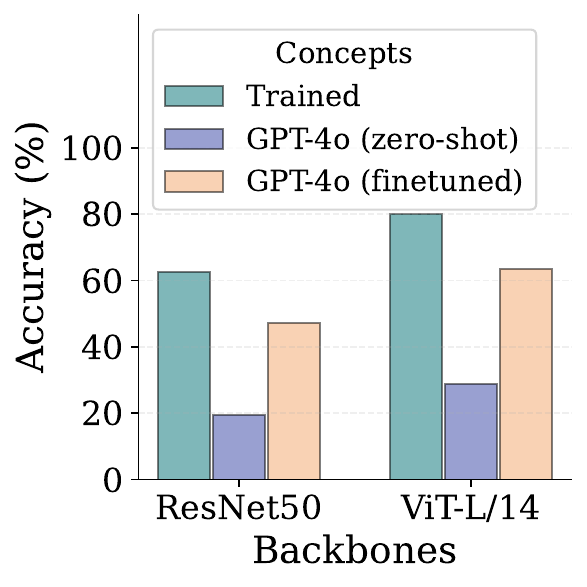}
    \caption{GPT, CUB}
\end{subfigure}
\hfill
\begin{subfigure}[t]{0.19\textwidth}
    \centering
    \includegraphics[width=\linewidth]{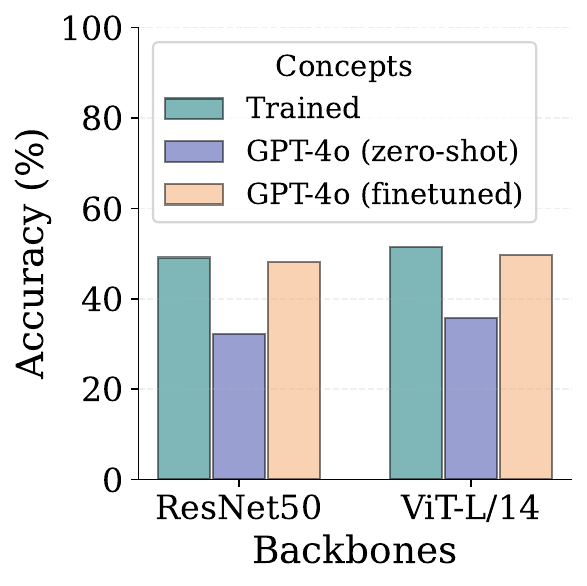}
    \caption{GPT, Places365}
\end{subfigure}
\hfill
\begin{subfigure}[t]{0.19\textwidth}
    \centering
    \includegraphics[width=\linewidth]{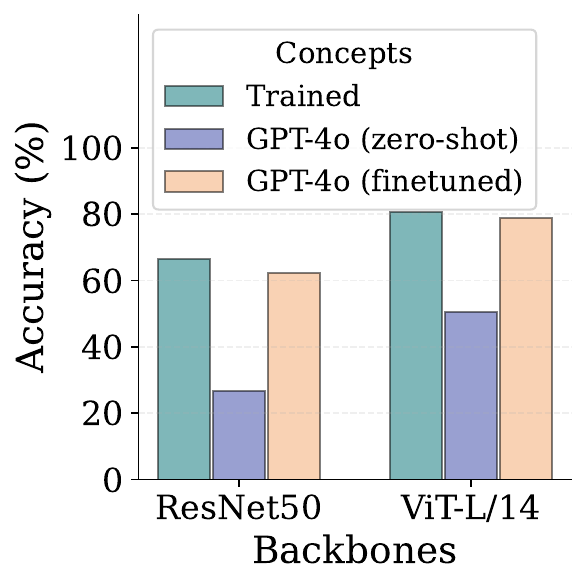}
    \caption{GPT, ImageNet}
\end{subfigure}

\caption{Adaptability of different concept pools across five datasets using ResNet50 and ViT-L/14 backbones. For each subfigure, we test the accuracy of FCBM using three types of concepts: the trained concepts, the LLM-generated concepts without training (zero-shot), and the LLM-generated concepts with only one epoch of fine-tuning (finetuned). DeepSeek-V3 (a-e) and GPT-4o (f-j) are employed as the LLM backbones, respectively.
}
\label{fig:zero_shot}
\end{figure*}

\begin{figure*}[tb!]
\centering
\includegraphics[width=16.2cm]{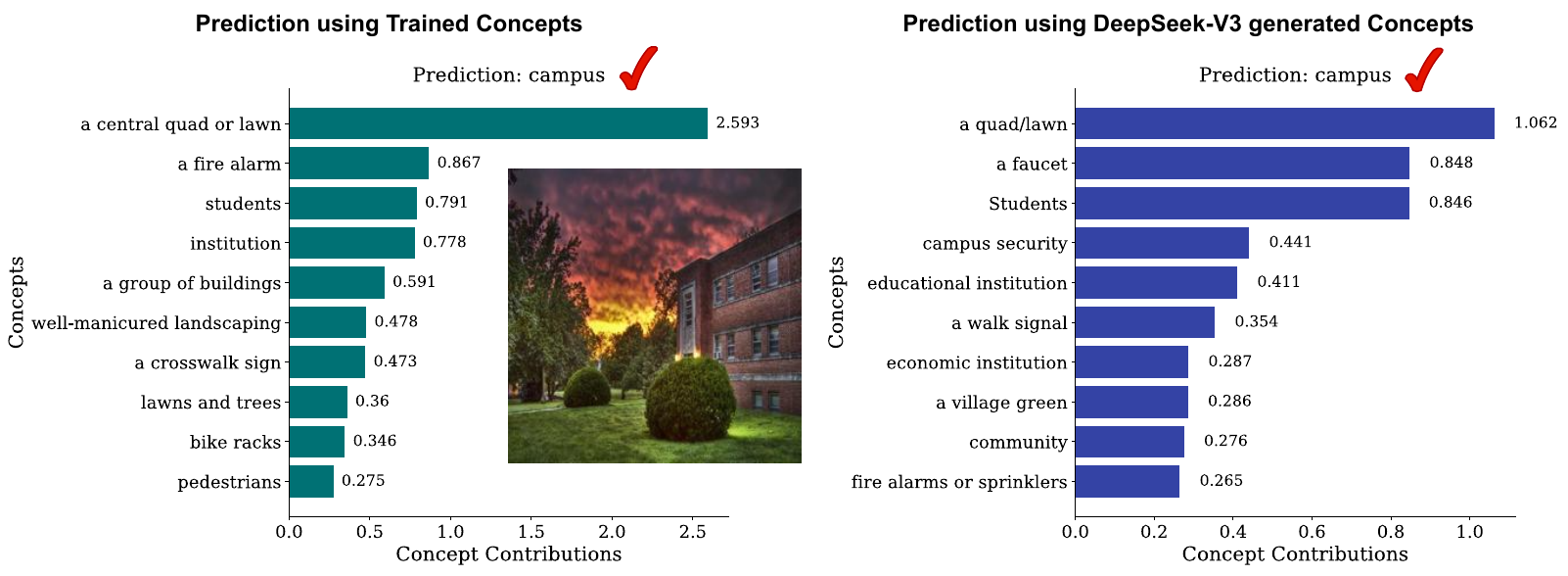}
\caption{
The adaptability of FCBM across different concept pools on the \textit{Places365} dataset. The left histograms illustrates the prediction made using the trained concepts, while the right histograms show the prediction based on the DeepSeek-V3-generated concepts. The left image belong to the ground-truth class, i.e., `campus'.
}
\label{fig:visualization}
\end{figure*}

\paragraph{Metrics.} The prediction accuracy is used as the main metric. To make a fairer comparison, we incorporate the average number of effective concepts (NEC), first introduced in~\citep{srivastava2024vlg} to keep similar sparsity of concept usages:
\begin{equation}
    \text{NEC}(W)=\frac{1}{n}\sum_{i=1}^m\sum_{j=1}^n \mathbb{I}\{W_{ij}\neq 0\}, 
\end{equation}
where $\mathbb{I}[\cdot]$ is an indicator returning $1/0$ with correct/wrong inputs. $W$ is the weights of the linear function in vanilla VLM-based CBMs~\citep{oikarinen2023labelfree}, and is $\mathring{h}(\bm{t})$ in our approach. In the main result, we set NEC near $30$, and we also conduct ablation studies in terms of NEC in Sec.~\ref{sec:ablation}. Experiments are repeated three times with different seeds, reporting the average performance with standard deviation. 

\paragraph{Baselines.} Six approaches are utilized as baselines, including the sparse and non-sparse standard model (linear projection from the concept layer to the prediction layer), PCBM~\citep{yuksekgonul2022post}, LF-CBM~\citep{oikarinen2023labelfree}, and CF-CBM~\citep{panousis2024coarse}. Note that all baselines are with similar NEC, except the non-sparse standard model.

\paragraph{Other Settings.} Other details are shown in the Appendix.

\subsection{Main Results}\label{sec:exp_main}

In this section, we compare the prediction accuracy of different approaches using two backbone models, i.e., ResNet50~\cite{he2016deep} and CLIP-ViT-L/14~\citep{radford2021learning}. The NEC for all approaches on each dataset is controlled to be approximately 30.

From a general perspective, the standard model achieves the highest accuracy across all datasets. This is expected, as it incorporates the most significant number of informative concepts for making predictions. However, its sparse version performs poorly, emphasizing the challenge of maintaining high accuracy while ensuring sparsity.

Among the other four baselines, the accuracy of PCBM~\citep{yuksekgonul2022post} is not particularly remarkable and is roughly comparable to that of the standard sparse model. Moreover, note that PCBM was not originally designed for larger datasets. In contrast, LF-CBM~\citep{oikarinen2023labelfree} and CF-CBM~\citep{panousis2024coarse} exhibit similarly strong accuracy with different design strategies, illustrating possible empirical upper bounds for accuracy when using sparse concepts.

Compared to these baselines, FCBM achieves the highest accuracy in most cases and ranks second in the remaining two, showing performance that matches or exceeds these state-of-the-art methods. This result suggests that replacing the linear function with a hypernetwork-sparsemax module does not degrade the final prediction quality. Beyond prediction accuracy, the next section will discuss zero-shot generalization and highlight the benefits brought by explicit concept learning.

\begin{figure*}[tb!]
\centering

\begin{subfigure}[t]{0.19\textwidth}
    \centering
    \includegraphics[width=\linewidth]{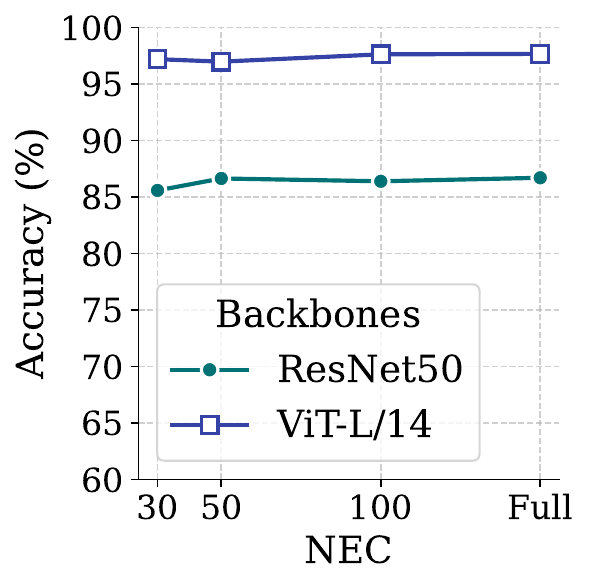}
    \caption{CIFAR10}
\end{subfigure}
\hfill
\begin{subfigure}[t]{0.19\textwidth}
    \centering
    \includegraphics[width=\linewidth]{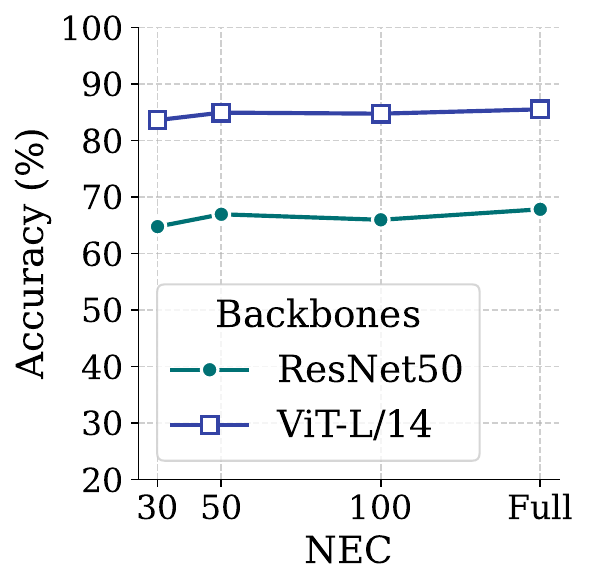}
    \caption{CIFAR100}
\end{subfigure}
\hfill
\begin{subfigure}[t]{0.19\textwidth}
    \centering
    \includegraphics[width=\linewidth]{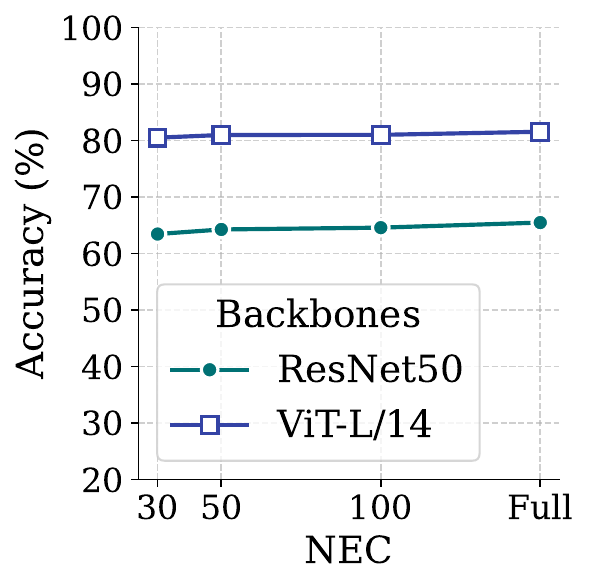}
    \caption{CUB}
\end{subfigure}
\hfill
\begin{subfigure}[t]{0.19\textwidth}
    \centering
    \includegraphics[width=\linewidth]{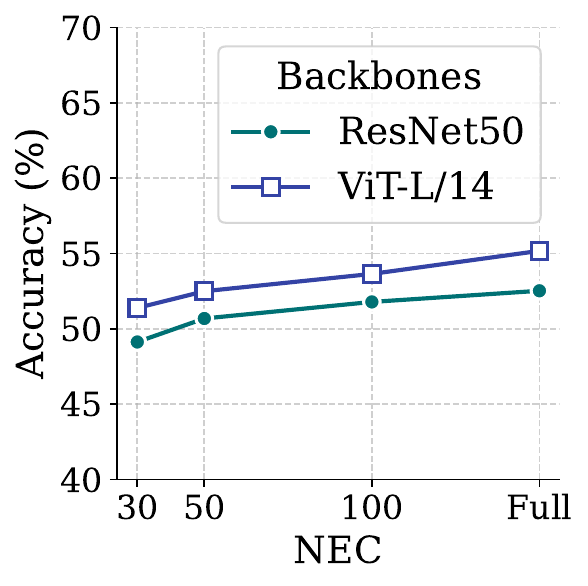}
    \caption{Places365}
\end{subfigure}
\hfill
\begin{subfigure}[t]{0.19\textwidth}
    \centering
    \includegraphics[width=\linewidth]{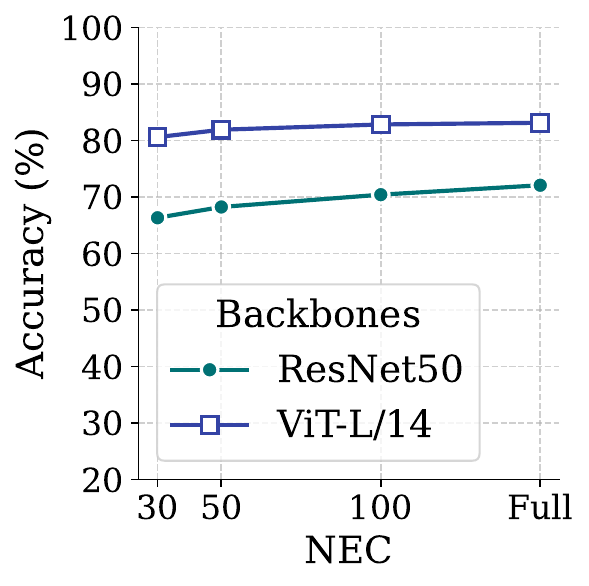}
    \caption{ImageNet}
\end{subfigure}

\caption{Sparsity analysis. We test the accuracy of FCBM with NEC = 30, 50, 100, and full concepts across five datasets using ResNet50 and ViT-L/14 backbones.}
\label{fig:sparsity}
\end{figure*}

\begin{table*}[tb!]
\centering
\scriptsize  
\begin{threeparttable}
\begin{tabular}{c|c|ccc|ccc|ccc|ccc|ccc}
\toprule
\multirow{2}{*}[-0.25em]{\textbf{Backbone}} & \multirow{2}{*}[-0.25em]{\textbf{Method}} 
& \multicolumn{3}{c|}{CIFAR10} 
& \multicolumn{3}{c|}{CIFAR100} 
& \multicolumn{3}{c|}{CUB} 
& \multicolumn{3}{c|}{Places365} 
& \multicolumn{3}{c}{ImageNet} \\
\cmidrule(lr){3-5} \cmidrule(lr){6-8} \cmidrule(lr){9-11} \cmidrule(lr){12-14} \cmidrule(lr){15-17}
& & Tr. & DS & GPT & Tr. & DS & GPT & Tr. & DS & GPT & Tr. & DS & GPT & Tr. & DS & GPT \\ \midrule

\multirow{3}{*}{ResNet50}
& Hard$^\#$ & 69.17 & 54.00 & 47.56 & 63.01 & 17.71 & 16.13 & 50.95 & 6.82 & 9.02 & \textbf{49.85} & 9.77 & 8.86 & 61.50 & 17.79 & 16.92 \\
& FCBM\textbackslash temp.$^\#$ & 81.27 & 62.62 & 59.37 & 49.80 & 29.13 & 31.89 & 29.45 & 10.97 & 12.38 & 33.95 & 22.98 & 23.61 & 39.60 & 17.95 & 16.60 \\
& FCBM & \textbf{85.59} & \textbf{75.32} & \textbf{75.09} & \textbf{64.77} & \textbf{35.57} & \textbf{34.54} & \textbf{63.92} & \textbf{18.07} & \textbf{19.43} & 49.13 & \textbf{32.79} & \textbf{32.27} & \textbf{66.34} & \textbf{29.23} & \textbf{26.69} \\ \midrule

\multirow{3}{*}{ViT-L/14}
& Hard$^\#$ & \textbf{97.27} & 78.78 & 73.10 & 65.15 & 23.61 & 24.68 & 68.15 & 9.14 & 12.87 & 50.37 & 11.80 & 11.60 & 75.22 & 15.07 & 14.36 \\
& FCBM\textbackslash temp.$^\#$ & 89.05 & 75.58 & 78.13 & 62.42 & 38.54 & 37.30 & 40.00 & 18.00 & 14.40 & 34.97 & 25.58 & 24.41 & 49.13 & 23.65 & 22.55 \\
& FCBM & 97.21 & \textbf{94.89} & \textbf{94.19} & \textbf{83.63} & \textbf{62.27} & \textbf{59.67} & \textbf{80.52} & \textbf{23.06} & \textbf{28.69} & \textbf{51.39} & \textbf{37.24} & \textbf{35.70} & \textbf{80.62} & \textbf{51.70} & \textbf{50.56} \\

\bottomrule
\end{tabular}
\begin{tablenotes}
\item[*] Tr. = Using trained concepts, DS = Using DeepSeek-V3-generated concepts, GPT = Using GPT-4o-generated concepts.
\item[$\#$] Hard: forcibly selecting a group of most effective concepts. FCBM\textbackslash temp.: FCBM using the default unlearnable temperature.
\end{tablenotes}
\end{threeparttable}
\caption{\label{tab:ablation}Accuracy of FCBM with different modules across five datasets using three groups of concepts. The best results for each backbone, dataset, and concept group are shown in \textbf{bold}.}
\end{table*}

\subsection{Adaptability Across Unseen Concept Pools}\label{sec:zero-shot}
To demonstrate FCBM's zero-shot generalization capability in terms of concepts, we follow the generation configuration proposed in~\citep{oikarinen2023labelfree} and regenerate the concepts using two widely used LLMs: DeepSeek-V3~\citep{liu2024deepseek} and GPT-4o~\citep{openai2024gpt4o}.

As shown in Fig.~\ref{fig:zero_shot}, FCBM demonstrates varying degrees of zero-shot generalization and fine-tuning adaptability across different concept pools. Our experimental setting is particularly challenging: we replace the entire concept pool with a new set of concepts, reflecting a realistic scenario in the era of rapidly evolving foundation models. Despite this difficulty, FCBM possesses a degree of zero-shot generalization, successfully transferring to new concepts without any fine-tuning, suggesting the learned structure within FCBM captures generalizable patterns in the semantic space. Remarkably, FCBM can adapt to the entirely new concept set with only a single epoch of fine-tuning, highlighting its ability to quickly incorporate evolving domain knowledge.

Fig.~\ref{fig:visualization} shows a example of concept contributions using the trained concepts (left) and the concepts generated by DeepSeek-V3 (right), where the image used for prediction belongs to the `campus' class in the \textit{Places365} dataset. Overall, the highly contributing concepts are similar, such as `a central quad or lawn' versus `a quad/lawn', `students' versus `Students', and `institution' versus `educational institution'. However, in the left prediction, the concept `a central quad or lawn' contributes the most, while in the right prediction, based on the DeepSeek-V3-generated concepts, the contributions are more dispersed. This dispersal may suggest that the concepts used in the right prediction do not fully align with those in the training set, which could potentially lead to classification errors. An additional case on the \textit{CUB} dataset can be seen in the Appendix.

To summarize, FCBM demonstrates the potential of zero-shot generalization in terms of concepts in VLM-based CBMs. However, we suggest that users exercise caution when using LLM-generated concepts for fine-grained categories or highly specialized datasets. 

\section{Ablation Studies}\label{sec:ablation}

\paragraph{Sparsity.} We evaluate the accuracy of FCBM with different numbers of effective concepts (NEC), including 30, 50, 100, and full concepts (all available concepts), across five datasets. As shown in Fig.~\ref{fig:sparsity}, the accuracy increases slightly as the number of NECs increases, but overall it remains stable. The most noticeable improvement is observed on the \textit{Places365} dataset, indicating the higher prediction difficulty for this dataset.

\paragraph{Sparsemax and Temperature.} In Table~\ref{tab:ablation}, we evaluate the accuracy of FCBM with three different modules: FCBM without sparsemax (denoted as `Hard', using hard truncation instead), FCBM without a learnable temperature (denoted as `FCBM\textbackslash temp.'), and the full version of FCBM. Two backbones (ResNet50 and ViT-L/14) and three groups of concepts (trained concepts, DeepSeek-V3 generated concepts, and GPT-4o generated concepts) are also included in this analysis. The hard truncation refers to the process of forcibly selecting the 30 most effective concepts (NEC).

From a general perspective, the full version of FCBM outperforms other methods in most cases, particularly in terms of zero-shot generalization performance. Although `Hard' occasionally achieves good results using trained concepts, its zero-shot generalization performance is the weakest, indicating that hard truncation does not effectively preserve concept understanding. Furthermore, `FCBM\textbackslash temp.' lacks the ability to control sparsity due to the absence of a learnable temperature. As a result, the sparsity is typically very low, making it difficult to select effective concepts.

\section{Conclusion and Outlook}

We proposed the Flexible Concept Bottleneck Model (FCBM), which addresses key challenges in CBMs by allowing dynamic adaptation of the concept pool. Through experiments on five datasets, FCBM demonstrated comparable performance to state-of-the-art methods, achieving competitive accuracy with a similar sparsity. Moreover, FCBM showed promising zero-shot generalization, successfully handling unseen concepts. Future work includes evaluating FCBM in more complex, real-world scenarios. It is interesting to investigate whether novel biomarkers can be seamlessly integrated into FCBM without sacrificing interpretability or predictive performance.

\clearpage

\medskip

\section*{Acknowledgement}
The work is supported by the National Natural Science Foundation of China (No. 62506367). R.Z. would like to acknowledge the supported by the China Postdoctoral Science Foundation under Grant Number 2025M771582 and the Postdoctoral Fellowship Program of CPSF under Grant Number GZB20250408. 

\bibliography{references}

\newpage
\appendix

\section{Sparsemax}\label{appdx:sparsemax}
\subsection{Derivation of Eq.~\ref{eq:jacabian}}
The original sparsemax function was proposed in \citep{martins2016softmax} with the Jacobian of the transformation:
\begin{equation}
\frac{\partial \mathcal{S}_{\max}^{\tau}(\bm{s})_i}{\partial \bm{s}_j} =
\begin{cases}
\delta_{ij} - \frac{1}{|P(\bm{s})|}, & \text{if } i,j \in P(\bm{s}),  \\
0, & \text{otherwise}, 
\end{cases}
\end{equation}
where $P(\bm{s})$ represents the support set (i.e., indices where $\mathcal{S}_{\max}^{\tau}(\bm{s})_i > 0$), and $\delta_{ij}$ is the Kronecker delta. Note that the Jacobian matrix of a transformation $\rho$ is $J_\rho (\bm{s}) := [\partial \rho_i(\bm{s})/\partial \bm{s}_j]$, so the full Jacobian matrix can be written as:
\begin{equation}
J_{\text{sparsemax}}(\bm{s}) = \text{Diag}(\bm{e}) - \frac{\bm{e}\bm{e}^\top}{|P(\bm{s})|}, 
\end{equation}
where $\bm{e}$ is an indicator vector with $s_i = 1$ if $i \in P(\bm{s})$ and $s_i = 0$ otherwise.

As noted in \citep{martins2016softmax}, for efficient computation during backpropagation, it is typically not necessary to compute the full Jacobian matrix, but only the product between the Jacobian and a given vector $\bm{v}$. For sparsemax, this product is given by:
\begin{equation}
J_{\text{sparsemax}}(\bm{s}) \cdot \bm{v} = \bm{e} \odot (\bm{v} - \hat{v}\mathbf{1}), \quad \text{with} \quad \hat{v} := \frac{\sum_{j \in P(\bm{s})} v_j}{|P(\bm{s})|}, 
\end{equation}
where $\odot$ denotes the Hadamard (element-wise) product and $\mathbf{1}$ is the all-ones vector. This computation is efficient, especially when the support set is small, requiring only $O(|P(\bm{s})|)$ operations.

\begin{table*}[t]
\centering
\begin{subtable}[t]{\textwidth}
\centering
\caption*{\textbf{(a) Accuracy (\%) across datasets}}
\footnotesize
\renewcommand{\arraystretch}{0.95}
\begin{tabular}{c|c|ccccc}
\toprule
\textbf{Backbone} & \textbf{Method} & CIFAR10 & CIFAR100 & CUB & Places365 & ImageNet \\ \midrule
\multirow{6}{*}{ResNet50}
& Standard & 88.55 {\scriptsize $\pm$ 0.02} & 70.19 {\scriptsize $\pm$ 0.06} & 71.00 {\scriptsize $\pm$ 0.16} & 53.28 {\scriptsize $\pm$ 0.04} & 73.14 {\scriptsize $\pm$ 0.04} \\ \cmidrule{2-7}
& Standard (sparse) & 82.11 {\scriptsize $\pm$ 0.00} & 57.54 {\scriptsize $\pm$ 0.01} & 53.34 {\scriptsize $\pm$ 0.01} & 44.00 {\scriptsize $\pm$ 0.00} & 57.03 {\scriptsize $\pm$ 0.01} \\
& PCBM & 76.43 {\scriptsize $\pm$ 0.02} & 56.24 {\scriptsize $\pm$ 0.06} & 58.39 {\scriptsize $\pm$ 0.06} & 42.26 {\scriptsize $\pm$ 0.35} & 61.38 {\scriptsize $\pm$ 0.22} \\
& LF-CBM & \textbf{86.16 {\scriptsize $\pm$ 0.05}} & \underline{64.62 {\scriptsize $\pm$ 0.06}} & 56.91 {\scriptsize $\pm$ 0.14} & \underline{48.88 {\scriptsize $\pm$ 0.05}} & \underline{66.03 {\scriptsize $\pm$ 0.05}} \\
& CF-CBM & 85.42 {\scriptsize $\pm$ 0.07} & 64.31 {\scriptsize $\pm$ 0.25} & \textbf{64.23 {\scriptsize $\pm$ 0.29}} & 46.39 {\scriptsize $\pm$ 0.34} & 65.95 {\scriptsize $\pm$ 0.27} \\
& FCBM (ours) & \underline{85.59 {\scriptsize $\pm$ 0.29}} & \textbf{64.77 {\scriptsize $\pm$ 0.41}} & \underline{63.46 {\scriptsize $\pm$ 0.41}} & \textbf{49.13 {\scriptsize $\pm$ 0.67}} & \textbf{66.34 {\scriptsize $\pm$ 0.34}} \\
\midrule
\multirow{6}{*}[-0.25em]{ViT-L/14}
& Standard & 98.02 {\scriptsize $\pm$ 0.03} & 86.99 {\scriptsize $\pm$ 0.09} & 85.22 {\scriptsize $\pm$ 0.05} & 55.66 {\scriptsize $\pm$ 0.06} & 84.11 {\scriptsize $\pm$ 0.05} \\ \cmidrule{2-7}
& Standard (sparse) & 96.61 {\scriptsize $\pm$ 0.00} & 80.28 {\scriptsize $\pm$ 0.00} & 72.07 {\scriptsize $\pm$ 0.00} & 44.90 {\scriptsize $\pm$ 0.01} & 72.96 {\scriptsize $\pm$ 0.13} \\
& PCBM & 92.13 {\scriptsize $\pm$ 0.05} & 74.29 {\scriptsize $\pm$ 0.13} & 68.25 {\scriptsize $\pm$ 0.32} & 43.18 {\scriptsize $\pm$ 0.36} & 65.79 {\scriptsize $\pm$ 0.53} \\
& LF-CBM & \underline{97.18 {\scriptsize $\pm$ 0.01}} & 81.98 {\scriptsize $\pm$ 0.02} & 75.44 {\scriptsize $\pm$ 0.04} & \underline{50.51 {\scriptsize $\pm$ 0.02}} & \underline{79.70 {\scriptsize $\pm$ 0.03}} \\
& CF-CBM & 96.35 {\scriptsize $\pm$ 0.05} & \underline{82.33 {\scriptsize $\pm$ 0.08}} & \underline{79.56 {\scriptsize $\pm$ 0.37}} & 48.55 {\scriptsize $\pm$ 0.28} & 79.16 {\scriptsize $\pm$ 0.44} \\
& FCBM (ours) & \textbf{97.21 {\scriptsize $\pm$ 0.06}} & \textbf{83.63 {\scriptsize $\pm$ 0.20}} & \textbf{80.52 {\scriptsize $\pm$ 0.27}} & \textbf{51.39 {\scriptsize $\pm$ 0.11}} & \textbf{80.62 {\scriptsize $\pm$ 0.32}} \\
\bottomrule
\end{tabular}
\end{subtable}

\begin{subtable}[t]{\textwidth}
\centering
\caption*{\textbf{(b) NEC across datasets}}
\footnotesize
\renewcommand{\arraystretch}{0.95}
\begin{tabular}{c|c|ccccc}
\toprule
\textbf{Backbone} & \textbf{Method} & CIFAR10 & CIFAR100 & CUB & Places365 & ImageNet \\ \midrule
\multirow{5}{*}{ResNet50}
& Standard (sparse) & 29.20 {\scriptsize $\pm$ 0.00} & 30.27 {\scriptsize $\pm$ 0.01} & 28.51 {\scriptsize $\pm$ 0.00} & 32.59 {\scriptsize $\pm$ 0.01} & 28.64 {\scriptsize $\pm$ 0.00} \\
& PCBM & 30.65 {\scriptsize $\pm$ 1.13} & 31.23 {\scriptsize $\pm$ 0.87} & 32.19 {\scriptsize $\pm$ 0.53} & 32.19 {\scriptsize $\pm$ 0.53} & 33.98 {\scriptsize $\pm$ 0.13} \\
& LF-CBM & 28.20 {\scriptsize $\pm$ 0.50} & 24.79 {\scriptsize $\pm$ 0.30} & 29.36 {\scriptsize $\pm$ 0.17} & 28.22 {\scriptsize $\pm$ 0.05} & 26.00 {\scriptsize $\pm$ 0.15} \\
& CF-CBM & 26.61 {\scriptsize $\pm$ 0.08} & 27.12 {\scriptsize $\pm$ 0.16} & 28.59 {\scriptsize $\pm$ 0.63} & 29.80 {\scriptsize $\pm$ 1.14} & 30.22 {\scriptsize $\pm$ 0.96} \\
& FCBM (ours) & 26.97 {\scriptsize $\pm$ 1.16} & 29.19 {\scriptsize $\pm$ 0.65} & 29.40 {\scriptsize $\pm$ 0.47} & 29.08 {\scriptsize $\pm$ 0.39} & 29.59 {\scriptsize $\pm$ 0.20} \\
\midrule
\multirow{5}{*}{ViT-L/14}
& Standard (sparse) & 27.17 {\scriptsize $\pm$ 0.12} & 29.39 {\scriptsize $\pm$ 0.00} & 27.54 {\scriptsize $\pm$ 0.04} & 31.24 {\scriptsize $\pm$ 0.01} & 29.26 {\scriptsize $\pm$ 0.00} \\
& PCBM & 30.20 {\scriptsize $\pm$ 0.56} & 30.84 {\scriptsize $\pm$ 0.73} & 33.40 {\scriptsize $\pm$ 0.46} & 34.29 {\scriptsize $\pm$ 0.09} & 34.05 {\scriptsize $\pm$ 0.12} \\
& LF-CBM & 25.60 {\scriptsize $\pm$ 0.21} & 27.26 {\scriptsize $\pm$ 0.02} & 27.87 {\scriptsize $\pm$ 0.09} & 29.01 {\scriptsize $\pm$ 0.09} & 29.52 {\scriptsize $\pm$ 0.02} \\
& CF-CBM & 26.87 {\scriptsize $\pm$ 0.11} & 27.44 {\scriptsize $\pm$ 0.13} & 28.52 {\scriptsize $\pm$ 0.40} & 29.63 {\scriptsize $\pm$ 1.06} & 30.42 {\scriptsize $\pm$ 1.18} \\
& FCBM (ours) & 23.23 {\scriptsize $\pm$ 1.04} & 28.75 {\scriptsize $\pm$ 0.27} & 29.33 {\scriptsize $\pm$ 0.12} & 29.66 {\scriptsize $\pm$ 0.22} & 29.42 {\scriptsize $\pm$ 0.66} \\
\bottomrule
\end{tabular}
\end{subtable}
\caption{\label{tab:appdx_sparsity}Prediction accuracy (\%) and NEC across five datasets using ResNet50 and ViT-L/14 backbones. The best and the second best results, excluding the `Standard' model, for each dataset and backbone are shown in \textbf{bold} and \underline{underline}, respectively.}
\end{table*}

\subsection{Derivation of Eq.~\ref{eq:approximation}}
For our temperature-controlled sparsemax, we need to derive the gradient with respect to the temperature parameter $\tau$. The relationship between the input and output of temperature-controlled sparsemax can be expressed as:
\begin{equation}
\tilde{\bm{s}}_i =
\begin{cases}
\bm{s}_i - \xi(\bm{s}) & \text{if } i \in P(\bm{s}),  \\
0 & \text{otherwise}, 
\end{cases}
\end{equation}
where $\xi(\bm{s})$ is the threshold function. Taking the derivative with respect to $\tau$ for elements in the support set:
\begin{equation}\label{eq:appdx_derivative}
\frac{\partial \tilde{\bm{s}}_i}{\partial \tau} = - \frac{\partial \xi(\bm{s})}{\partial \tau}.
\end{equation}
The threshold function $\xi(\bm{s})$ with temperature can be expressed as:
\begin{equation}
\xi(\bm{s}) = \frac{\left(\sum_{j \in P(\bm{s})} \bm{s}_j\right) - \tau}{|P(\bm{s})|} = \frac{\sum_{j \in P(\bm{s})} \bm{s}_j}{|P(\bm{s})|} - \frac{\tau}{|P(\bm{s})|}.
\end{equation}
Taking the derivative with respect to $\tau$:
\begin{equation}
\frac{\partial \xi(\bm{s})}{\partial \tau} = - \frac{1}{|P(\bm{s})|}.
\end{equation}
Pubstituting this back to Eq.~\ref{eq:appdx_derivative}:
\begin{equation}
\frac{\partial \tilde{\bm{s}}_i}{\partial \tau} = \frac{1}{|P(\bm{s})|}.
\end{equation}

Then, The overall derivation of $\tau$ is formulated as
\begin{equation}
\frac{\partial \mathcal{L}}{\partial \tau} = \sum_{i \in P(\mathbf{\bm{s}})} \frac{\partial \mathcal{L}}{\partial \tilde{\bm{s}}_i} \cdot \frac{\partial \tilde{\bm{s}}_i}{\tau} = \sum_{i \in P(\mathbf{\bm{s}})} \frac{1}{|P(\bm{s})|}\cdot \frac{\partial \mathcal{L}}{\partial \tilde{\bm{s}}_i}.
\end{equation}

\section{Experiments}\label{appdx:exp}

\subsection{Detailed Sparsity in Table~\ref{tab:main}}
We detail the NEC across five datasets along with accuracy in Table~\ref{tab:appdx_sparsity}.

\subsection{Further Experimental Protocols}\label{appdx:setting}

\noindent \textbf{Temperature Decay.} During the training of FCBM, the temperature is initially decayed exponentially, which allows the model to first learn quickly from dense concepts and then focus on the sparsity of concepts. The decay rate is empirically set to 0.998 for small-scale datasets (CIFAR10, CIFAR100, and CUB) and 0.92 for large-scale datasets by grid search. Once the NEC falls below a specified threshold (e.g., NEC $\textless$ 30 in main results), the decay process stops, and the temperature is continuously optimized according to Eq.~\ref{eq:approximation}.

\noindent \textbf{Backbone Usage.} In the experiments, we use ResNet50 and CLIP-ViT-L/14 backbones for the target models, while the CLIP models used are CLIP-ViT-B/16 and CLIP-ViT-L/14, respectively. The former choice aligns with the settings of existing baselines (although some baselines use CUB-pretrained ResNet18 for the CUB dataset to obtain higher accuracy), while the latter demonstrates the performance improvements achievable with more powerful backbones.
The pretrained CLIP models (ResNet50, ViT-B/32, and ViT-L/14) follow the official implementations provided by OpenAI~\footnote{\url{openaipublic.azureedge.net}}. For all models, the image embeddings are taken from the final visual encoder layer, and the text embeddings are taken from the final transformer layer corresponding to the $\texttt{<EOT>}$ token.

\noindent \textbf{Structure of Hypernetwork.} The hypernetwork is a three-layer perceptron with 4096 neurons in each hidden layer. A ReLU activation function is applied between each pair of layers to improve generalization. The module concludes with an $l_2$-normalization.

\noindent \textbf{LLM Usage.} We use DeepSeek-V3~\citep{liu2024deepseek}, and GPT-4o~\citep{openai2024gpt4o} to generate concepts, respectively, in Sec.~\ref{sec:zero-shot}. The versions used are as follows:
\begin{itemize}
    \item \textbf{DeepSeek-V3}: DeepSeek-V3-2024-12-26.
    \item \textbf{GPT-4o}: GPT-4o-2024-11-20.
\end{itemize}

\begin{figure*}[tb!]
\centering

\includegraphics[width=\textwidth]{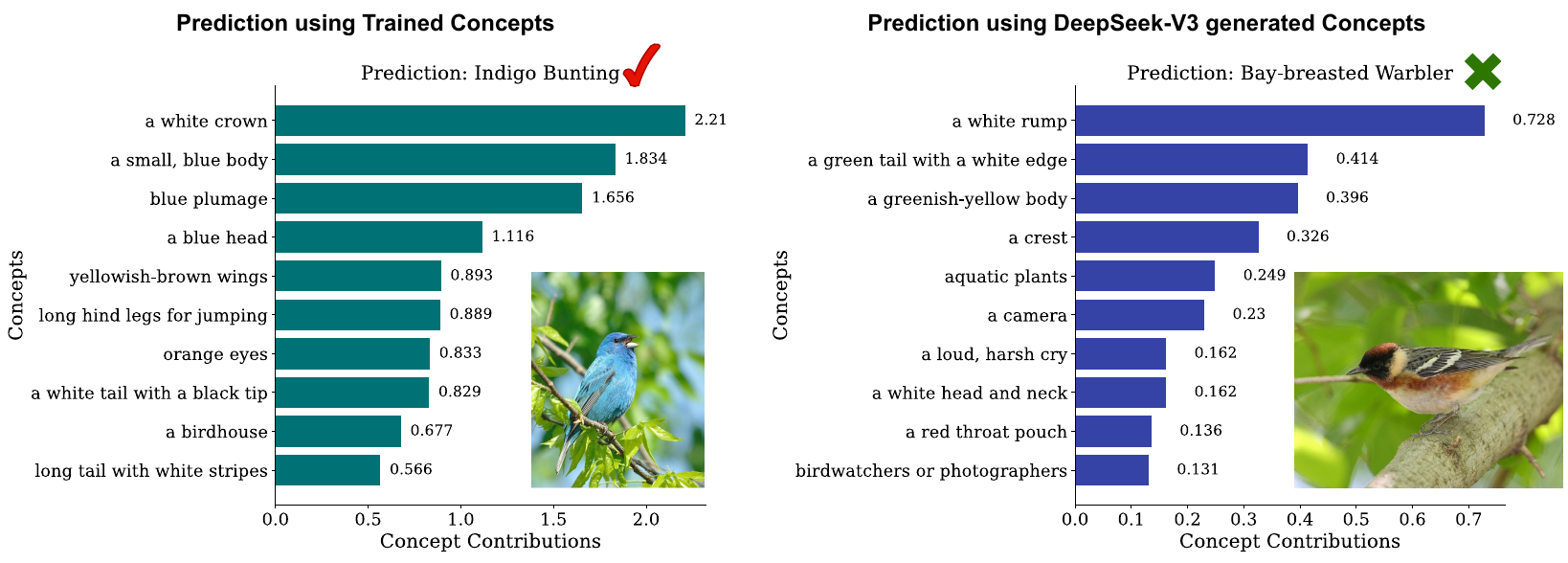}
\caption{\label{fig:histogram2} An incorrect zero-shot prediction on CUB. The left histograms illustrates the prediction made using the trained concepts, while the right histograms show the prediction based on the DeepSeek-V3-generated concepts. The left image belong to the ground-truth class ‘Indigo Bunting’, while the right image in illustrates the characteristics of the class ‘Bay-breasted Warbler’.
}
\end{figure*}

\noindent \textbf{Additional Case on \textit{CUB} Dataset.}
Fig.~\ref{fig:histogram2} shows further differences between the two predictions, where the image used for prediction belongs to the `Indigo Bunting' class in the \textit{CUB} dataset. Intuitively, the correct prediction is primarily driven by the `blue'-related concepts. However, we observe that \textbf{`blue' is not present} in the entire concept group generated by DeepSeek-V3. As a result, the model relies on other concepts for prediction, which ultimately leads to an incorrect classification. The differences in concept selection and contributions may thus help explain the observed variations in prediction accuracy.

\end{document}